\documentclass{article}
\usepackage{ijcai21}

\usepackage{times}

\usepackage{soul}
\usepackage{url}
\usepackage[hidelinks]{hyperref}
\usepackage[utf8]{inputenc}
\usepackage[small]{caption}
\usepackage{graphicx}
\usepackage{amsmath,amssymb}
\usepackage{booktabs}
\usepackage[noend]{algpseudocode}
\usepackage{algorithm}
\usepackage{hyperref}       
\usepackage{url}            
\usepackage{booktabs}       
\usepackage{amsfonts}       
\usepackage{nicefrac}       
\usepackage{microtype}      
\usepackage{xcolor}
\usepackage{amsfonts}
\usepackage{mathtools}
\newcommand\myeq{\stackrel{\mathclap{\normalfont\mbox{def}}}{=}}
\usepackage[export]{adjustbox}
\usepackage{graphicx,subfigure}
\usepackage{wrapfig,lipsum,booktabs}
\usepackage{amsmath}

\DeclareMathOperator*{\argmin}{arg\,min}

\title{Knowledge Consolidation based Class Incremental\\ Online Learning with Limited Data}

\author{
Mohammed Asad Karim$^1$\footnote{Contact Author}\and
Vinay Kumar Verma$^2$\and
Pravendra Singh$^3$\and \\
Vinay Namboodiri$^{1,4}$\And
Piyush Rai$^1$\\
\affiliations
$^1$Indian Institute of Technology Kanpur, India\\
$^2$Duke University, United States\\
$^3$Indian Institute of Technology Roorkee, India\\
$^4$University of Bath, United Kingdom\\
\emails
asadkarim0938@gmail.com,
vinaykumar.verma@duke.edu,
pravendra.singh@cs.iitr.ac.in,\\
vpn22@bath.ac.uk,
piyush@cse.iitk.ac.in
}

\begin{document}

\maketitle

\begin{abstract}
We propose a novel approach for class incremental online learning in a limited data setting. This problem setting is challenging because of the following constraints: (1) Classes are given incrementally, which necessitates a class incremental learning approach; (2)  Data for each class is given in an online fashion, i.e., each training example is seen only once during training; (3) Each class has very few training examples; and (4) We do not use or assume access to any replay/memory to store data from previous classes. Therefore, in this setting, we have to handle twofold problems of catastrophic forgetting and overfitting. In our approach, we learn robust representations that are generalizable across tasks without suffering from the problems of catastrophic forgetting and overfitting to accommodate future classes with limited samples. Our proposed method leverages the meta-learning framework with knowledge consolidation. The meta-learning framework helps the model for rapid learning when samples appear in an online fashion. Simultaneously, knowledge consolidation helps to learn a robust representation against forgetting under online updates to facilitate future learning. Our approach significantly outperforms other methods on several benchmarks.
\end{abstract}

\section{Introduction}
\label{sec:intro}
Deep neural networks have achieved promising results on various tasks.
However, these models suffer from the problem of catastrophic forgetting~\cite{kirkpatrick2017overcoming}. The most prominent reason for catastrophic forgetting is that the model is not trained to also remember the previous knowledge when acquiring new knowledge.  In general, the model is trained to optimize its performance on the current task with no consideration of how the updated model will perform on earlier tasks. This greedy update overwrites the parameter values that may have been optimal for previous tasks. \emph{Continual-learning (CL)} (also sometimes referred to as lifelong/incremental learning) is a learning paradigm to address this issue in deep neural networks and has been gaining significant attention in recent work~\cite{Parisi2019}. 

We present a novel approach for \emph{class incremental} online learning problem in a  limited data setting. This problem setting is more challenging than standard class incremental learning~\cite{javed2019meta} due to additional constraints: (1) Data in each class appears in the online fashion, i.e., the model sees every training example exactly once; (2) The number of training examples in each class is very small; and (3) We do not use any replay/memory to store the training examples from previous classes. This is the most general setting for class incremental learning and various practical usage scenario can be obtained through this or a relaxed setting. For instance, in face recognition, it is common to have few examples per class but usually not in an online learning fashion, whereas for a robot navigating in an environment, the setting would also be online. We empirically show that learning a robust representation that can accommodate future tasks may be a potential solution to handle the problem mentioned above. Our proposed approach achieves this by leveraging the meta-learning~\cite{finn2017model} framework with knowledge consolidation.

Meta-learning~\cite{finn2017model} has proven to be an effective approach for learning generic feature representations that can be rapidly adapted to new tasks by fine-tuning using very few examples (and in some cases, even \emph{without} fine-tuning~\cite{vinyals2016matching,raghu2019rapid}). While such use of meta-learning might seem appealing and does indeed show some promising results in continual learning settings~\cite{javed2019meta}, in practice, this approach is still prone to the problem of catastrophic forgetting. One of the reasons for this is the \emph{overparametrized} nature of deep neural networks, in which only a few neurons are activated/fired for all samples. As a result, the network is reliant only on a small set of parameters. Although this may not be a problem when learning only a single task, it can potentially be an issue in continual learning where we are required to learn a sequence of tasks and, while learning a new task, any changes to these parameters can drastically affect the performance on the older tasks.

\begin{figure*}[!htbp]
    \centering
    \includegraphics[scale=0.4]{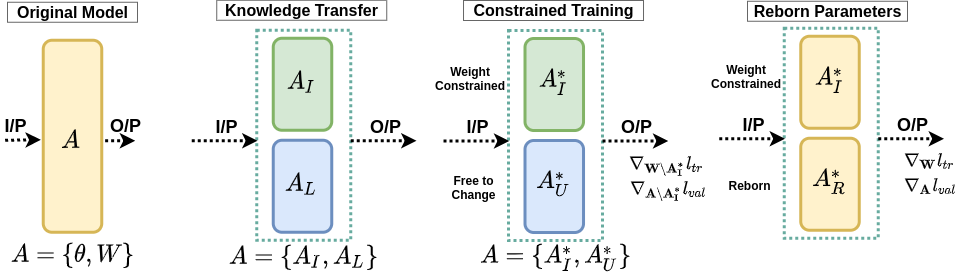}
    \caption{The figure shows the various steps in knowledge consolidation. The original model is learned via a meta-learner. Thereafter, we transfer the model's knowledge into a subset of parameters and partition the model into important and unimportant parameters. Then we retrain the model by allowing $\mathbf{A}_{U}^*$ to change freely according to the loss function and also constraining $\mathbf{A}_{I}^*$ to preserve previous knowledge.}
    \label{fig:my_label}
\end{figure*}

To address this, we present a \emph{knowledge consolidation} based meta-learning approach. In our approach, during training, we identify and split the network parameters into two groups, called important and unimportant. The network's existing knowledge is squeezed into the set of important parameters, and the unimportant/dead parameters are freed, thereby expanding the network's capacity to accommodate the future learning trajectories. Briefly, a learning trajectory is a sequence of examples where examples from a particular class occur together in the sequence (we discuss this in detail later). The proposed strategy ensures that the knowledge from the old learning trajectories is preserved in a compressed form within a small set of important network parameters, which we identify and isolate, and \emph{then} move on to adopt new learning trajectories. The extra knowledge obtained via learning from the new trajectories updates the unimportant parameters, and they also become important. In knowledge consolidation, we rejuvenate the dead neurons in the model and consolidate the knowledge of the previously preserved parameters and the new rejuvenated parameters. This helps to learn a robust representation. Therefore, the model capacity is fully utilized, and small changes in a few parameters strongly resist the catastrophic forgetting. The knowledge consolidation process overview is shown in Fig.~\ref{fig:my_label}.

Note that we use knowledge consolidation and meta-learning in the training phase. Training is done on a base class set using multiple learning trajectories. We strictly follow class incremental online learning setting within a learning trajectory. However, across learning trajectories, the class incremental learning setting is not used (since multiple learning trajectories can have the same set of classes). Therefore, we do not follow the incremental learning setting during training on the base class set (since training is done on multiple learning trajectories). We follow this to make the model's representation robust and facilitate future continual learning during evaluation time (Section~\ref{sec:probform}), where we perform class incremental online learning with limited data on a \emph{novel} class set. We use entirely different (disjoint) classes in the novel class set than the base class set, and during evaluation on the novel class set, we only use meta-learning for quick adaptation. 

Our approach significantly outperforms other incremental learning methods by a significant margin. We show that a basic online updating strategy on representations learned by our approach Knowledge Consolidation based Class Incremental Online Learning (KCCIOL) is better than memory-based rehearsal methods. Our approach can also be integrated with existing continual learning approaches such as MER, EWC,  ER-Reservoir as shown in Section~\ref{rreclcomp}.

 \section{Problem Formulation and Evaluation Protocol for Novel Class Set Testing}\label{sec:probform}
Let \{$\tau_1,\tau_2 \dots \tau_k \dots \tau_l \dots$\} denote a stream of learning-trajectories and $\forall i$, $\tau_i \sim P_{test}(\tau)$ where $P_{test}(\tau)$ denotes the trajectory distribution during testing from the novel class set.
Each learning-trajectory $\tau_i$ is further split into two sets -- train and validation, i.e., $\tau_i=\{\tau_{tr},\tau_{val}\}$ where  $\tau_{tr}=\{x_n,y_n\}_{n=1}^k$ and $\tau_{val}=\{x_n,y_n\}_{n=k+1}^{k+s}$ are the labeled samples, $\{k,s\}\in \mathbb{N}$. Here, $\forall n$, $(x_n,y_n)\in (\mathcal{X}_{test},\mathcal{Y}_{test})$ denote the input and label pairs and the trajectory distribution $P_{test}(\tau)$ is defined over $(\mathcal{X}_{test},\mathcal{Y}_{test})$ which is disjoint from  base training class set $(\mathcal{X}_{train},\mathcal{Y}_{train})$ i.e. $\mathcal{Y}_{train}\cap \mathcal{Y}_{test}=\varnothing$. Moreover, we assume $class(\tau_{tr}) =  class(\tau_{val})$ i.e. classes of $\tau_{tr}$ are the same as classes of  $\tau_{val}$. 
The goal of continual learning is to minimize the loss on the unseen examples of classes learnt earlier in an incremental fashion, and can be written as:
$\mathbb{E}_{\tau \sim P_{test}(\tau)}[\mathcal{L}(f(\tau_{val}^x|\mathbf{\theta},\mathbf{W}),\tau_{val}^y)]$ i.e model is evaluated on $\tau_{val}$.
Our evaluation protocol (Algorithm 4) is similar to the class-incremental setting~\cite{javed2019meta}, but samples within each class also arrive in an online manner. In particular, here are the key differences: 1) We assume availability of very few samples per class; 2) For a particular class, each sample is seen exactly once; and 3) We do not use any replay mechanism. These differences make our problem setting considerably more challenging than the standard class-incremental setting.

\section{Training on Base Class Set using Meta Learning Approach}
\label{sec:metacont}

Following a similar notation as in Sec.~\ref{sec:probform}, let \{$\tau_1,\tau_2 \dots \tau_k \dots \tau_l \dots$\} denote a stream of learning trajectories and $\forall i$, $\tau_i \sim P_{train}(\tau)$ where $P_{train}(\tau)$ denotes the learning trajectory distribution during training. Following the model-agnostic meta learning (MAML) set-up~\cite{finn2017model}, we assume that the $i^{th}$ learning-trajectory's 
data $\tau_i$ is further split into two sets, meta-train and meta-val, i.e. $\tau_i=\{\tau_{tr},\tau_{val}\}$ where  $\tau_{tr}=\{x_n,y_n\}_{n=1}^k$ and $\tau_{val}=\{x_n,y_n\}_{n=k+1}^{k+s}$ are the labeled samples, $\{k,s\}\in \mathbb{N}$. Here, $\forall n$, $(x_n,y_n)\in (\mathcal{X}_{train},\mathcal{Y}_{train})$ denote the input and label pairs and the learning trajectory distribution $P_{train}(\tau)$ is defined over $(\mathcal{X}_{train},\mathcal{Y}_{train})$. Moreover, we assume $class(\tau_{tr})\subset class(\tau_{val})$ i.e. classes of $\tau_{tr}$ are a proper subset $\tau_{val}$; therefore $\tau_{val}$ contains all the classes of $\tau_{tr}$ and some additional classes. Note that $\tau_{val}$ and $\tau_{tr}$ are the random trajectories of length $k$ and $s$, respectively. In $\tau_{tr}$, samples of each class occur together, i.e samples from class 2 occur after and before samples from class 1 and class 3, respectively. We sample a learning trajectory multiple times for training on a base class set. We randomly sample a learning trajectory by selecting a  subset (proper) of classes (randomly) from the base class set. Therefore, every learning trajectory has a different class order. 

We follow the MAML setting of continual learning~\cite{javed2019meta}. In the inner loop, each class arrives in a sequential manner, and the boundary of each class is available in advance. In the outer-loop, the aim is to minimize the empirical risk over the unseen data, provided the model is optimized on the seen data in an online fashion with the continual learning constraint i.e. when learning from the training data of current class, we are not allowed to access the training data of previous classes. The overall loss function across all trajectories is defined as: 
\begin{equation}
    \mathbb{E}_{\tau \sim P_{train}(\tau)}[\mathcal{L}(f(\tau_{val}^x|\mathbf{\theta},\mathbf{W}),\tau_{val}^y)]
\end{equation}
$\mathcal{L}(f(\tau_{val}^x|\mathbf{\theta},\mathbf{W}),\tau_{val}^y)$ denotes the loss on the model $f$ for the 
validation trajectory of $\tau$ . For notational brevity (and slight abuse of notation), we use $\tau_{val}^x$ to refer to all the inputs of validation trajectory of task $\tau$ and $\tau_{val}^y$ to refer to the corresponding true labels. 

The function $f:\mathcal{X}\rightarrow \mathcal{Y}$ is defined as $f(\tau^x|\mathbf{\theta},\mathbf{W})=g(h(\tau^x|\mathbf{\theta})|\mathbf{W})$ where $h_{\mathbf{\theta}}:\mathcal{X}\rightarrow \mathbb{R}^d$ is defined by parameter $\mathbf{\theta}$ (representation learning parameters) and $g_{\mathbf{W}}:\mathbb{R}^d\rightarrow\mathcal{Y}$ is defined by $\mathbf{W}$. The classifier parameters $\mathbf{W}$ are learned using meta-train set ($\tau_{tr}$), and representation learning parameters $\mathbf{\theta}$ and $\mathbf{W}$ are jointly learned using the meta-val set ($\tau_{val}$).

In the inner-loop of the meta-learner, which learns $\mathbf{W}$, the model is trained on the meta-train data $\tau_{tr}$. The outer loop, which learns $\theta$ and $\mathbf{W}$ is trained using meta-val data $\tau_{val}$. In the outer loop, the model's loss is also computed on novel classes not seen during the inner-loop training since the classes in $\tau_{tr}$ are a proper subset of classes in $\tau_{val}$. Evaluation on both sets $\tau_{tr}$ and $\tau_{val}$ makes the model perform well on both current and previously learned classes. The optimization problems solved by the inner loop and the outer loop are given by:
\begin{equation}\label{eq:2}
    \small
    \mathbf{W} = \argmin_{\mathbf{W}}l_{tr}(\mathbf{\theta},\mathbf{W})\myeq\mathcal{L}(f(\tau_{tr}^x|\mathbf{\theta},\mathbf{W}),\tau_{tr}^y)
\end{equation}
\begin{equation}\label{eq:3}
    \small
    (\theta,\mathbf{W}) = \argmin_{\mathbf{\theta},\mathbf{W}}l_{val}(\mathbf{\theta},\mathbf{W})\myeq\mathcal{L}(f(\tau_{val}^x|\mathbf{\theta},\mathbf{W}),\tau_{val}^y)
\end{equation}

In Eq.~\ref{eq:2}, for notational simplicity, we use the entire training data but during training we perform an online update over $\tau_{tr}$. The above two optimization problems are solved in an alternating fashion using $\tau_{tr}$ and $\tau_{val}$, respectively, with the most recent parameter $\mathbf{W}$ obtained from Eq.~\ref{eq:2} used in Eq.~\ref{eq:3}.

\section{Knowledge Consolidation based Meta Learning for Training on Base Class Set}
\label{sec:constrained}

While the meta-learning based approach for continual learning described in the above section is promising, it is not particularly effective for our problem setting. One of the reasons for this is the \emph{overparametrized} nature of deep neural networks, in which only a few neurons are activated/fired for all samples. As a result, the network is reliant only on a small set of parameters. Although this may not be a problem for a single task learning setting, it can potentially be an issue in continual learning where we are required to learn a sequence of tasks and, while learning a new task, any changes to these parameters can drastically affect the performance on the older tasks. We overcome this by using a knowledge consolidation based meta-learning approach. Our proposed approach identifies the important and unimportant/dead parameters, rejuvenates the dead parameters, and consolidates the knowledge of the important and reborn parameters (Fig.~\ref{fig:my_label}). Therefore, the model capacity is fully utilized, and small changes in a few parameters strongly resist the catastrophic forgetting.

A simple way to assess the importance of a parameter is to use its absolute value \cite{han2015learning}, often used in deep model compression. We can discard weights/parameters having small absolute values without sacrificing upon the model's performance~\cite{han2015learning}. We leverage this simple idea to identify the important parameters in the model effectively.  
The proposed approached modifies the meta-learning framework by introducing knowledge consolidation. We define $\mathbf{A}=[\mathbf{\theta},\mathbf{W}]$ as the joint set of parameters of the complete model. We partition model parameters $\mathbf{A}$ into two disjoint sets. The important parameters are denoted by $\mathbf{A}_I$ and the ``less important'' ones are denoted by $\mathbf{A}_L$, s.t., $\mathbf{A}= \{\mathbf{A}_I, \mathbf{A}_L\}$ and $\mathbf{A}_I\cap \mathbf{A}_L=\varnothing$.

 Most of the model's knowledge is contained in $\mathbf{A}_I$. Our goal is to \emph{preserve} the knowledge present in $\mathbf{A}_I$. We apply a weight-constrained regularization on $\mathbf{A}_I$ to ensure minimal changes when a new trajectory is learned. On the other hand, we let $\mathbf{A}_L$ be free to change in order to accommodate new trajectories. Therefore,  while learning a new set of  trajectories, the following regularized loss function is optimized:
\begin{equation}\label{eq:7}
    \sum_{\tau \sim P_{train}({\tau})}  \mathcal{L}(f(\tau_{val}^x|\mathbf{\theta},\mathbf{W}),\tau_{val}^y)+\mathcal{R}(\mathbf{A}_I)
\end{equation}
One way to define the weight-constrained regularization $\mathcal{R}(\mathbf{A_I})$ would be $\lambda||\mathbf{A}_I^{t+1}-\mathbf{A}_I^t||_F$, where $\mathbf{A}_I^t$ is the important weights after  $t^{th}$ step. The large value of $\lambda$ ensures minimal changes in the important weights. 

Na\"ively partitioning the model into $\{\mathbf{A}_I, \mathbf{A}_L\}$ (based on absolute value) often does not show any significant improvement since various techniques like dropout and batch normalization force the model's knowledge to be shared across all model parameters, which causes $\mathbf{A}_L$ to contain non-negligible knowledge. Ideally, the value of unimportant weights should be zero. However, in reality, this is not the case and the set $\mathbf{A}_L$ usually contains the non-negligible information. Therefore, we first distill the model's knowledge into a subset of parameters $\mathbf{A}_I^{*}$ (important parameters set) such that the remaining part  $\mathbf{A}_U^{*}$ (unimportant parameters set) contains negligible information. The weight-constrained regularization can now be imposed on $\mathbf{A}_I^{*}$, while the set $\mathbf{A}_U^{*}$ is free to be adapted for the new trajectories. To transfer/distill the model's knowledge to a subset of the parameters, we \emph{finetune} the complete model with the following $\ell_1$ regularized objective:
\begin{equation}\label{eq:8}
  \sum_{\tau \sim P_{train}({\tau})}  \mathcal{L}(f(\tau_{val}^x|\mathbf{\theta},\mathbf{W}),\tau_{val}^y)+\gamma||\mathbf{A}||_1
\end{equation}
We can maintain model performance by using an appropriate hyperparameter $\gamma$. The $\ell_1$ regularizer forces the model knowledge to be squeezed in a subset of model parameters $\mathbf{A}_I^{*}$. Rest of the model parameters $\mathbf{A}_U^{*}$ contain negligible information and  therefore are free to change. Now, the set  $\mathbf{A}$ can be split into important parameters $\mathbf{A}_I^{*}$ and unimportant parameters set $\mathbf{A}_U^{*}$, i.e., $\mathbf{A}= \{\mathbf{A}_I^{*}, \mathbf{A}_U^{*}\}$ and $\mathbf{A}_I^{*}\cap \mathbf{A}_U^{*}=\emptyset$. Given this updated set of important and unimportant/dead set of parameters, the outer-loop optimization of the meta-learner is given by (akin to Eq-\ref{eq:7})
\begin{equation}
\small
    \begin{split}
        (\theta,\mathbf{W}) = \argmin_{\mathbf{\theta},\mathbf{W}}l_{val}(\mathbf{\theta},\mathbf{W}) & \myeq\mathcal{L}(f(\tau_{val}^x|\mathbf{\theta},\mathbf{W}),\tau_{val}^y) \\ + \lambda||{\mathbf{A}_I^{*^{t}}}-\mathbf{A}_I^{*^{t+1}}||_F
    \end{split}
\end{equation}
\begin{equation}
    \mathbf{A}_I^{*^{t+1}}-\mathbf{A}_I^{*^{t}} \approx \nabla_{\mathbf{A}_I^{*^{t}}}(\mathcal{L}(f(\tau_{val}^x|\mathbf{\theta},\mathbf{W}),\tau_{val}^y))
\end{equation}
To preserve the knowledge contained in  $\mathbf{A}_I^{*}$, we apply weight-constrained regularization on $\mathbf{A}_I^{*}$ as above, which ensures that $\mathbf{A}_I^{*}$ do not change drastically when new learning-trajectories are encountered. Rest of the parameters ($\mathbf{A}_U^{*}$) are free to change. Therefore, we ``rejuvenate'' the parameters in $\mathbf{A}_U^{*}$. Representing these rejuvenated set of parameters as $\mathbf{A}_R^{*}$, now the consolidated knowledge from both $\mathbf{A}_R^{*}$ and $\mathbf{A}_I^{*}$ provides a robust representation for our problem setting. Therefore, small changes in a few parameters strongly resist catastrophic forgetting because the model capacity is fully utilized, and model predictions are not reliant on a small set of parameters. As demonstrated by our experiments, such a parameter rejuvenation and knowledge consolidation significantly enhance the performance of a meta-learner based model for class incremental online learning. For a summarized algorithmic description of our approach, please refer to the Algorithms 1, 2, 3, 4.

\section{Related Work}

\textbf{Incremental Learning Methods:}
Rehearsal based incremental learning methods \cite{rebuffi2017icarl,shin2017continual,chaudhry2018efficient,isele2018selective,rolnick2019experience,chaudhry2020using} store a part of training data and re-train on this stored data while training on new tasks. Regularization based incremental learning methods \cite{lopez2017gradient,kirkpatrick2017overcoming,chaudhry2018efficient} add an additional regularization term in the loss function, which prevents the weights from changing drastically when moving from one task to the next task. DER \cite{buzzega2020dark} uses both rehearsal and regularization. PODNet \cite{douillard2020podnet} uses a spatial distillation-loss along with a representation made of proxy vectors from each class. The dynamic network methods  \cite{singh2020calibrating,singh2021rectification}  are also proposed for incremental learning.

\begin{algorithm}
\caption{Training Algorithm}
\textbf{Require:} $\gamma$: Learning coefficient for L1 loss\\
\textbf{Require:} $\lambda$: Learning coefficient for constraint loss\\
\textbf{Require:} $\delta$: Fraction of model parameters ($\tfrac{|A_{I}^{*}|}{|A|}$)\\
\textbf{Require:} $\{\alpha_{i},\beta_{i}\}_{i=1}^{3}$: Inner and outer loop lr (learning rate) 
\textbf{Require:} $\{steps_{i}\}_{i=1}^{3}$: No of steps for KCCIOL algorithm 
\begin{algorithmic}[1]
\State Randomly Initialize model parameters $\Theta$, 
\State $mask$ = 0
\State $\Theta$  = $\mathbf{KCCIOL}(\alpha_{1}, \beta_{1}, 0,0,mask,\Theta, steps_{1})$
\State $\Theta$ = $\mathbf{KCCIOL}(\alpha_{2}, \beta_{2}, 0,\gamma, mask, \Theta, steps_{2})$
\State $mask  = \mathbf{GetMask}(\Theta,\delta)$
\State $\Theta = \mathbf{KCCIOL}(\alpha_{3}, \beta_{3}, \lambda,0, mask, \Theta, steps_{3})$
\end{algorithmic}
\end{algorithm}

\begin{algorithm}

\caption{KCCIOL}

\textbf{Require :} $p_{train}(\tau)$: Distribution over learning trajectories \\
\textbf{Require :} $mask$: Index matrix  \\
\textbf{Require :} $\Theta$: Model parameters
\begin{algorithmic}[1]
\For {$i=1$, . . , $ steps$}
\State Sample learning trajectory $\tau_{i} \sim  p_{train}(\tau)$
\State $\{\tau_{tr}, \tau_{val}\}  = \tau_{i}$
\State $\{\theta, W$\} = $\Theta$
\State $W_{0} = W$
\For {$j$=1, 2 , . . , $k$}
\State $(X_{j},Y_{j}) = (\tau_{tr}^{x}[j], \tau_{tr}^{y}[j]) $

\State $W_{j} = W_{j - 1} - \alpha\nabla_{W_{j - 1}}[\mathcal{L}(f(X_{j}|\mathbf{\theta},W_{j-1}),Y_{j})]$
\EndFor

\State $l_{meta} =[\mathcal{L}(f(\tau_{val}^x|\mathbf{\theta},W_{k}),\tau_{val}^y)] $

\State $l_{constraint} = || mask*\nabla_{\theta,W}l_{meta}||_{2}^{2}$
\State $l_{1} = ||\Theta||_{1} $
\State Update $\Theta \leftarrow \Theta - \beta \nabla_{\theta, W}(l_{meta}+\lambda l_{constraint}+ \gamma l_{1})$

\EndFor
\State \textbf{return} $\Theta$
\end{algorithmic}
\end{algorithm}

\begin{algorithm}
\caption{Mask Calculation}
\textbf{Require:}  $\Theta$: Model Parameters \\
\textbf{Require:} $\delta$: Fraction of model parameters ($\tfrac{|A_{I}^{*}|}{|A|}$)
\begin{algorithmic}[1]
\Procedure{getMask}{$\Theta$, $\delta$}
 \State $threshold  = percentile(|\Theta|, 1-\delta)$
 \State $mask = zeros(len(\Theta))$
 \State $index = 0$
 \While{$index <= len(\Theta$)}
 \If{$mask[index]>=threshold$}
 \State $mask[index] = 1$
 \State $index++$
 \EndIf
 \EndWhile
 \State \textbf{return} $mask$
\EndProcedure
\end{algorithmic}
\end{algorithm}

\begin{algorithm}
\caption{Evaluation Protocol}

\textbf{Require :} $p_{test}(\tau)$: Distribution over learning trajectories \\
\textbf{Require :} $\theta$: Representation Learning Parameters
\begin{algorithmic}[1]

\State Randomly Initialize $W$
\State Sample learning trajectory $\tau_{i} \sim  p_{test}(\tau)$
\State $\{\tau_{tr}, \tau_{val}\}  = \tau_{i}$

\For {$j=$1, . . , $k$}
\State $(X_{j},Y_{j}) = (\tau_{tr}^{x}[j], \tau_{tr}^{y}[j]) $
\State $W_{j} = W_{j - 1} - \alpha\nabla_{W_{j - 1}}[\mathcal{L}(f(X_{j}|\mathbf{\theta},W_{j-1}),Y_{j})]$
\EndFor

\State \textbf{return} $Accuracy(f(\tau_{val}^x|\mathbf{\theta},W_{k}),\tau_{val}^y)$
\end{algorithmic}
\end{algorithm}

\textbf{Meta-Learning Methods:}
MER \cite{riemer2018learning} uses meta-learning to learn parameters that prevent interference and enable knowledge transfer based on future gradients. OML \cite{javed2019meta} is a meta-learning approach that focuses on learning a generic representation that prevents catastrophic forgetting during online learning.  OSAKA \cite{caccia2020online} is a general approach to continual learning where the agent must be able to solve previously unseen distribution tasks with minimal forgetting on previous tasks. OSAKA is different from the standard incremental learning setting as seen tasks can be revisited, and online average accuracy is reported at the end of the training instead of reporting accuracy on all seen tasks. Therefore, it would be unfair to compare our method with OSAKA. 

\textbf{Online Continual Learning Methods:}
MERLIN \cite{joseph2020meta} is a replay based method for online continual learning which learns a meta-distribution from which task specific parameters are sampled at the time of inference. \cite{von2019continual} proposed a method similar to MERLIN where deterministic task specific weights are generated using hypernetworks \cite{ha2016hypernetworks}.  GSS  uses constrained optimization for sampling samples for replay instead of random sampling for online continual learning. Incremental learning in online scenario \cite{He_2020_CVPR} tackles the problem of catastrophic forgetting in the online scenario under different setting than ours. \cite{He_2020_CVPR} is a memory based method where future data consists of samples from new classes as well as unseen samples from old classes, whereas, in our approach, data from one class are seen together. CTN \cite{phamcontextual} is a bi-level optimization network that uses a context network to model task-specific features which address catastrophic forgetting and knowledge transfer. But the context network needs task-specific knowledge and semantic memory to function, whereas our method does use replay and is task-free.

\section{Experiments}
We evaluate our approach via extensive experiments across various datasets. We follow the evaluation protocol where the model is updated in an online fashion and later evaluated on the unseen data (Section~\ref{sec:probform}). We compare the performance of our model (KCCIOL) against several baselines. 
 
 \textbf{Baselines:} In the \textit{Scratch} baseline, we evaluate the performance of a randomly initialized network. In \textit{Pretrained} baseline, the network is pretrained on the train set. The \textit{SRNN} approach uses a Set-KL method proposed by \cite{liu2019utility} to learn a sparse representation using train set. \textit{MRCL} is a recent approach originally proposed by \cite{javed2019meta} to train a model in a meta-learning setup for continual learning tasks. \textit{OML} is a modified version of MRCL proposed by \cite{javed2019meta} where the classifier parameters are randomly re-initialized at each step of the training. \textit{MAML-Rep} is also a MAML \cite{finn2017model} based algorithm similar to OML and MRCL, where batch updates are performed in the inner-loop. 
 
  \begin{table}[t]
    \centering
    \scalebox{0.67}{
    \addtolength{\tabcolsep}{-4.4pt}
    \begin{tabular}{cccccccccc}
        \toprule
         \textbf{Classes} &\textbf{Scratch} & \textbf{Pretrained} &\textbf{SRNN} & \textbf{MRCL} &\textbf{MAML-Rep} &\textbf{OML}   & \textbf{Ours}\\
        \midrule
        10 & 15.9 $\pm$ 3.5 & 42.6 $\pm$ 10 & 70.4 & 83.8 $\pm$ 6.2 & 86.1 &92.6 $\pm$ 3.5   & \textbf{95.4 $\pm$ 3.4}\\
        
        50 & 2.4 $\pm$ 1.4 & 24.4$\pm$ 4.3 &53.9 & 66.5 $\pm$ 4.0 & 71.3 &81.3 $\pm$ 2.4   & \textbf{85.8 $\pm$ 2.5}\\
         
        100 & 1.5 $\pm$ 0.3 & 15.5 $\pm$ 1.9 & 44.3 &51.8 $\pm$ 2.6& 70.0 &76.1 $\pm$ 2.0  & \textbf{81.4 $\pm$ 2.3}\\
       
        150 & 1.2 $\pm$ 0.5 & 11.7 $\pm$ 1.1 &27.0 & 42.8 $\pm$ 2.5& 53.0 &65.2 $\pm$ 2.2  & \textbf{77.1 $\pm$ 1.6}\\
       
        200 & 0.8  $\pm$ 0.5 & 8.0 $\pm$ 1.1 &18.3 & 33.8 $\pm$ 1.8& 35.7 &59.3 $\pm$ 1.8  & \textbf{72.6 $\pm$ 1.5}\\
       \bottomrule
        
        \end{tabular}}\\
    \caption{Classification accuracy (mean$\pm$std) on the omniglot dataset averaged across 50 test trajectories randomly sampled from the meta-test set. Classes column refers to the total number of classes in the sampled trajectory.}
    \label{tab:Table1}
\end{table}

\subsection{Experiments on Omniglot Dataset:}
\noindent\textbf{Implementation Details}: The Omniglot dataset \cite{lake2015human} contains 1623 classes of different handwritten characters from 50 different alphabets.  Each class contains 20 samples with 15/5 as the train/test split.  
The first 963 classes constitute the $(\mathcal{X}_{train}, \mathcal{Y}_{train})$ and the remaining classes are used as $(\mathcal{X}_{test}, \mathcal{Y}_{test})$. For learning trajectory during training, $\tau_{tr}$ consists of 10 samples from a class randomly sampled from the training set. $\tau_{val}$ consists of 10+1 samples where ten samples are randomly sampled from the train set, and the 11th sample belongs to the class used in $\tau_{tr}$. During the evaluation, each learning trajectory consists of an equal number of classes in both $\tau_{tr}$ and $\tau_{val}$ which are sampled from the test set. We use 15/5 samples per class for $\tau_{tr}$/$\tau_{val}$ during evaluation. We use Adam optimizer, and the first six layers are used for learning representation.

\noindent\textbf{Hyperparameter Settings:} We train our model using hyperparameters: $\beta_{1}$ = 1e-4, $\alpha_{1} = $ 1e-2, $steps_{1}$ = 20000, $\beta_{2}$ = 1e-4, $\alpha_{2} = $ 1e-2 , $\gamma = $5e-5, $steps_{2}$ = 15000, $\beta_{3}$ = 1e-4, $\alpha_{3} = $ 1e-2 , $\lambda = $5e-4, $steps_{3}$ = 4000, $\delta$ = 0.5.

\noindent\textbf{Model Architecture:}
We use six convolutional layers followed by two fully connected layers, and each convolutional layer contains 256 filters of 3 $\times$ 3 kernel size with (2, 1, 2, 1, 2, 2) strides (same as used in \cite{javed2019meta}). ReLU activation function is used for the non-linearity.

The proposed approach KCCIOL performs better by a significant margin on a wide range of classes as compared to the other baselines. We achieve a significant performance boost of 13.3 \%, when 200 classes are learnt continually as shown in Table~\ref{tab:Table1}. 

\subsection{Experiments on Mini-Imagenet Dataset}

\noindent\textbf{Implementation Details:}
Vinyals et al. \cite{vinyals2016matching} proposed the mini-imagenet dataset, which is a subset of the imagenet dataset. There are a total of 100 classes with 600 colored images of size $84 \times 84$. We use 64 classes for training and 20 classes for testing. For learning trajectory during training, $\tau_{tr}$ consists of 10 samples from a class randomly sampled from the training set. $\tau_{val}$ consists of 15 samples where 10 samples are randomly sampled from the training set, and 5 samples belong to the class used in $\tau_{tr}$. During the evaluation, we sample classes from the test set for creating a learning trajectory. We  use 30 samples per class for $\tau_{tr}$ and $\tau_{val}$. We use Adam optimizer, and the first six layers are used for learning representations.

\noindent\textbf{Hyperparameter Settings:}
We train our model using hyperparameters: $\beta_{1}$ = 1e-4, $\alpha_{1} = $ 1e-1, $steps_{1}$ = 26000, $\beta_{2}$ = 1e-4, $\alpha_{2} = $ 1e-1 , $\gamma = $1e-4, $steps_{2}$ = 26000, $\beta_{3}$ = 1e-4, $\alpha_{3} = $ 0.5 , $\lambda = $1e-4, $steps_{3}$ = 22000, $\delta$ = 0.5.

\noindent\textbf{Model Architecture:}
The model architecture for the Mini-Imagenet is same as used in the Omniglot dataset.

\begin{table}[t]
    \centering
    \scalebox{0.75}{
    \addtolength{\tabcolsep}{-3pt}
    \begin{tabular}{ccccccc}
         \toprule
         \textbf{Classes}& \textbf{Scratch}& \textbf{Pretrained} & \textbf{SRNN} &\textbf{MRCL}&\textbf{OML}  & \textbf{Ours}\\
        \midrule
        6 & 19.3 $\pm$ 4.0 & 16.9 & 34.2 & 41.8 $\pm$ 9.0 & 37.0 $\pm$ 6.8 & \textbf{49.4 $\pm$ 6.5}  \\
        
        8 & 12.9 $\pm$ 1.3 & 13.0&29.2 & 36.8 $\pm$ 7.1 & 34.6 $\pm$ 4.2 & \textbf{45.3 $\pm$ 4.9}\\
        
        10 & 10.1 $\pm$ 0.9& 10.0& 25.8 & 34.4 $\pm$ 6.5 & 33.2 $\pm$ 4.6  & \textbf{42.7 $\pm$ 4.6}\\
       
        12 & 8.5 $\pm$ 0.9 & 8.5 & 23.1  &30.6 $\pm$ 4.9 & 29.5 $\pm$ 3.2 & \textbf{39.4 $\pm$ 3.0}\\
        
        14 & 9.1 $\pm$ 1.9 & 8.8 & 20.4 &29.6 $\pm$ 4.5 & 27.8 $\pm$ 2.5 & \textbf{37.0 $\pm$ 3.7}\\
        
        16 & 7.4 $\pm$ 1.8 & 10.8 & 18.1 &28.0 $\pm$ 3.5 & 25.2 $\pm$ 2.0 & \textbf{33.4 $\pm$ 2.1} \\
        
        18 & 5.5 $\pm$ 0.7 & 10.4 & 17.7  & 26.3 $\pm$ 3.3 & 23.9 $\pm$ 2.1 & \textbf{31.9 $\pm$ 2.2}\\
       
        20 & 6.2 $\pm$ 1.4 & 10.0 & 17.3 &25.5 $\pm$ 2.8 & 22.9 $\pm$ 1.6 & \textbf{29.7 $\pm$ 1.4} \\
       \bottomrule
        \end{tabular}}\\
    \caption{Classification accuracy (mean$\pm$std) on mini-imagenet dataset averaged across 50 test trajectories randomly sampled from the meta-test set. Classes column refers to the total number of classes in the sampled trajectory.}
    \label{tab:Table2}
\end{table}

 Our method KCCIOL consistently outperforms others as shown in  Table~\ref{tab:Table2}. We get a performance boost upto  12.4 \%  over the OML.

\begin{table}[t]
    \centering
\scalebox{0.75}{
    \addtolength{\tabcolsep}{-2pt}
    \begin{tabular}{lcccc}
        \toprule
        
        \textbf{Method}  &\textbf{Standard} & \textbf{Pre-Training} &\textbf{OML} &\textbf{Ours}  \\
        \midrule
        
        Online  & 4.64 $\pm$ 2.61 & 21.16 $\pm$ 2.71 & 64.72 $\pm$ 2.57 & \textbf{85.68 $\pm$ 2.10}  \\
        Approx. IID  & 53.95 $\pm$ 5.50 & 54.29 $\pm$ 3.48 & 75.12 $\pm$ 3.24 & \textbf{88.66 $\pm$ 2.10}  \\
        MER  & 54.88 $\pm$ 4.12 &62.76 $\pm$ 2.16 & 76.00 $\pm$ 2.07 &  \textbf{91.28 $\pm$ 1.38}  \\
        EWC  & 5.08 $\pm$ 2.47 &18.72 $\pm$ 3.97 & 64.44 $\pm$ 3.13 &  \textbf{87.10 $\pm$ 1.40}  \\
        ER-Reservoir  & 52.56 $\pm$ 2.12 &36.72 $\pm$ 3.06 & 68.16 $\pm$ 3.12 &  \textbf{90.10 $\pm$ 1.35}  \\
        \bottomrule
        \end{tabular}}\\
    \caption{KCCIOL combined with existing continual learning methods on the omniglot dataset. We use 50 tasks with 1 class per task. The accuracies are averaged over 10 runs.}
    \label{tab:Table5}
\end{table}

\begin{table}[t]
    \centering
\scalebox{0.75}{
    \addtolength{\tabcolsep}{-2pt}
    \begin{tabular}{lcccc}
        \toprule
        
        \textbf{Method}  &\textbf{Standard} & \textbf{Pre-Training} &\textbf{OML} &\textbf{Ours}  \\
        \midrule
        
        Online  & 1.40 $\pm$ 0.43 & 11.80 $\pm$ 1.92 & 55.32 $\pm$ 2.25 & \textbf{80.10 $\pm$ 1.71}  \\
        Approx. IID  & 48.02 $\pm$ 5.67 &46.02$\pm$ 2.83 & 67.03 $\pm$ 2.10 &  \textbf{85.90 $\pm$ 1.76}  \\
        MER  & 29.02 $\pm$ 4.01 &42.05$\pm$ 3.71 & 62.05 $\pm$ 2.19 &  \textbf{83.42 $\pm$ 1.67}  \\
        EWC  & 2.04 $\pm$ 0.35 &10.03 $\pm$ 1.53 & 56.03 $\pm$ 3.20 &  \textbf{82.90 $\pm$ 1.27}  \\
        ER-Reservoir  & 24.32 $\pm$ 5.37 &37.44 $\pm$ 1.67 & 60.92 $\pm$ 2.41 &  \textbf{84.76 $\pm$ 1.12}  \\
        \bottomrule
        \end{tabular}}\\
    \caption{KCCIOL combined with existing continual learning methods on omniglot. We use 100 tasks with 5 classes per task. The acccuracies are averaged over 10 runs. }
    \label{tab:Table6}
\end{table}

\begin{table}[!t]
    \centering
    \scalebox{0.6}{
    \addtolength{\tabcolsep}{-3pt}
    \begin{tabular}{ccccccccc}
         \toprule
         \textbf{Classes}  & \textbf{DER} & \textbf{DER++} & \textbf{HAL}  & \textbf{MERLIN} & \textbf{GSS} & \textbf{PODNet} & \textbf{Ours}\\
        \midrule
        6   & 27.22 $\pm$ 6.39 & 36.77$\pm$5.73 & 26.33 $\pm$ 3.44  & 22.78  & 35.00 & 42.68 $\pm$ 2.33 & \textbf{49.4 $\pm$ 5.5}\\
        
        8   & 22.29 $\pm$ 5.42 & {28.13 $\pm$ 3.46} & 20.63 $\pm$ 3.11  & 14.16  & 30.00 & 36.10 $\pm$ 2.00 & \textbf{45.3 $\pm$ 4.9}\\
        
        10   & 19.0 $\pm$ 2.69 & {23.5 $\pm$ 4.14} & 15.93 $\pm$ 4.14  & 11.24  & 24.33 & 31.48 $\pm$ 1.56 & \textbf{42.7 $\pm$ 4.6}\\
       
        12   & 15.58 $\pm$ 3.35 & {19.14 $\pm$ 3.51} & 14.86 $\pm$ 2.45  & 9.42 & 17.78 & 27.94 $\pm$ 1.42 & \textbf{39.4 $\pm$ 3.0}\\
        
        14   & 13.93 $\pm$ 2.18 & {17.14 $\pm$ 3.27} & 12.36 $\pm$ 1.58  & 7.53 & 17.14 & 25.15 $\pm$ 1.29 & \textbf{37.0 $\pm$ 3.7}\\
        
        16   & 13.33 $\pm$ 2.83 & {14.21 $\pm$ 2.83} & 10.75 $\pm$ 1.56  & 6.83 & 11.87 & 22.93 $\pm$ 1.29 &  \textbf{33.4 $\pm$ 2.1}\\
        
        18   & 11.48 $\pm$ 1.97 & {13.15 $\pm$ 2.96} & 9.46 $\pm$ 1.55  & 6.16 & 12.41 & 21.14 $\pm$ 1.22 &  \textbf{31.9 $\pm$ 2.2}\\
       
        20  & 9.68 $\pm$ 3.07 & {12.48 $\pm$ 2.08}  & 8.9 $\pm$ 1.50  & 6.11 & 13.00 & 19.63 $\pm$ 1.14 & \textbf{29.7 $\pm$ 1.4}\\
       \bottomrule
        \end{tabular}}\\
    \caption{Performance comparison with recent SOTA incremental learning methods on the mini-imagenet dataset. Classification accuracy (mean$\pm$std) is averaged across 50 test trajectories randomly sampled from the meta-test set. Classes column refers to the total number of classes in the sampled trajectory.}
    \label{tab:sota}
\end{table}

\section{KCCIOL Complements Existing Continual Learning Methods}
\label{rreclcomp}
We have demonstrated the efficiency of KCCIOL when learning-trajectories are learnt in a continual online fashion. In this section, we will demonstrate that the representation generated by KCCIOL not only performs well on online updates but it also greatly improves the performance of other continual learning methods such as MER \cite{rolnick2019experience}, EWC \cite{kirkpatrick2017overcoming}, ER-Reservoir \cite{rolnick2019experience} when they use our model as base network. Online, Approx IID baselines have been taken from \cite{javed2019meta}.

We evaluate the model's performance under four different settings for each continual learning method, which are Standard, Pre-Training, OML, and KCCIOL. In OML and KCCIOL, we use the $\theta$ generated by these methods as the base of the model and do not update the $\theta$ parameters. In the Pre-Training setting, we train the model independently on the training data and again keep the $\theta$ fixed during the process. The difference between the Pre-Training and Standard setting is that all the parameters of the model are updated in the case of the Standard setting, while in the pre-training, only $\mathbf{W}$ are updated. In the standard setting, the model is pretrained independently on train data to prevent other settings from having an unfair advantage of being trained on the training set. From Tables-\ref{tab:Table5}, \ref{tab:Table6} we can observe that KCCIOL with just online updates significantly outperforms other continual learning methods even when they are combined with OML. KCCIOL can do so because it can learn a generic robust feature representation that is generalizable across all tasks, enabling the model to perform future learning. For the train/test split, we use 15/5 samples per class. Even without using any memory/replay, KCCIOL can outperform the replay based methods by a significant margin.

\section{Performance Comparison with Recent SOTA Incremental Learning Methods}

We also compare our method KCCIOL with the recent state-of-the-art incremental learning methods on the mini-imagenet dataset under incremental online learning setting. From Table~\ref{tab:sota}, we can observe that recent incremental learning methods such as DER \cite{buzzega2020dark}, PODNet \cite{douillard2020podnet}, HAL \cite{chaudhry2020using} perform poorly in an incremental online setting.  Even SOTA methods for online learning, such as  GSS \cite{aljundi2019gradient}, MERLIN \cite{joseph2020meta} suffer from catastrophic forgetting when training data is limited. We only use thirty samples per class for experiments on the mini-imagenet dataset.  We observe that KCCIOL outperforms the next best method by more than 10\% when ten classes are learnt continually. Also, all of the above-mentioned methods are replay based approaches, while KCCIOL does not use any replay/memory.

\section{Conclusion}
We propose a novel approach to learn robust feature representations that are generalizable across future learning classes. Our approach uses a meta-learning framework with knowledge consolidation for learning generic feature representations that can be rapidly adapted for future classes without forgetting the previous classes under online updates to facilitate future learning. Our approach shows a significant improvement for class incremental online learning on several benchmark datasets.
\section*{Acknowledgements}
PR thanks support from Qualcomm Innovation Fellowship and Visvesvaraya Young Faculty Fellowship.
\newpage
\small
\bibliographystyle{named}
\bibliography{egbib}


\section{Supplementary Material}

\subsection{Experiments on CIFAR-100 Dataset} 
\noindent\textbf{Implementation Details:}
CIFAR-100 \cite{krizhevsky2009learning} is a labeled subset of the 80 million tiny images dataset. There are a total of 100 classes and 600 images of 32 $\times$ 32 per class. The first 70 classes are used for training, and the last 30 classes are used as a test set. We use the same split for learning trajectories for training and testing as used in the case of imagenet.

\noindent \textbf{Hyperparameter Settings:}
We train our mode using hyperparameters: $\beta_{1}$ = 1e-4, $\alpha_{1} = $ 1e-3, $steps_{1}$ = 50000, $\beta_{2}$ = 1e-4, $\alpha_{2} = $ 1e-3 , $\gamma = $1e-4, $steps_{2}$ = 35000, $\beta_{3}$ = 1e-4, $\alpha_{3} = $ 0.1 , $\lambda = $5e-4, $steps_{3}$ = 21000, $\delta$ = 0.5.

\noindent\textbf{Model Architecture:}
We use five convolutional layers followed by two fully connected layers with 256 convolutional filters of 3 $\times$ 3 kernel size with strides [1, 2, 1, 2, 1].

We use the first 70 classes as a training set and the last 30 classes as the test set. From Table~\ref{tab:Table3}, we observe an absolute improvement of  23.9 \% over the close competitor MRCL, when 10 classes are learned continually. 

\begin{table}[h]
    \centering
\scalebox{0.7}{
    \addtolength{\tabcolsep}{7pt}
    \begin{tabular}{ccccc}
        \toprule
         \textbf{Classes}  &\textbf{Scratch} &\textbf{MRCL} &\textbf{Ours}\\
        \midrule
        10  & 10.3$\pm$1.3 & 20.0 $\pm$ 4.3 & \textbf{43.9 $\pm$ 6.0}\\
       
        15  & 7.0 $\pm$0.8 &16.6 $\pm$ 2.5 & \textbf{37.5 $\pm$ 4.0}\\
       
        20  & 5.1$\pm$ 0.3 &13.8 $\pm$ 2.4  & \textbf{32.7 $\pm$ 2.8}\\
        
        25  & 4.2$\pm$ 0.8 &13.4 $\pm$ 1.7   & \textbf{29.3 $\pm$ 1.5}\\
        
        30  & 3.4 $\pm$ 0.1  &11.6 $\pm$ 2.3  & \textbf{26.0 $\pm$ 3.1}\\
        \bottomrule
        \end{tabular}}\\
        \caption{Results on CIFAR-100 dataset for the different number of classes incrementally learned in an online fashion.}
    \label{tab:Table3}
\end{table}

\subsection{Experiments on Regression Tasks} 
\noindent\textbf{Implementation Details:}
We follow the same setting as \cite{javed2019meta}. We use sine functions for our experiments. We create functions by randomly sampling amplitude, phase from [0.1, 5] and [0, $\pi$] respectively. 400 randomly sampled sine functions are used as the training set, and 500 functions are used as the testing set. A learning trajectory is created by randomly selecting ten functions from the train/test set for training/evaluation. We use 1280 samples for $\tau_{tr}$ from each of the ten functions. We use 32 samples for $\tau_{val}$ from each of the 10 functions. Each sample in the learning trajectory could be denoted by $(X = (n,z),Y)$ where $Y = \sin_{n}(z)$, $n \in [1,10]$ . We perform batch-online updates over the training data of the learning trajectory with batch size 32. We use Adam optimizer, and the first six layers are used for learning representations.

\noindent \textbf{Hyperparameter Settings:}
We train our mode using hyperparameters: $\beta_{1}$ = 1e-4, $\alpha_{1} = $ 3e-3, $steps_{1}$ = 20000, $\beta_{2}$ = 2.7e-6, $\alpha_{2} = $ 3e-3 , $\gamma = $1e-5, $steps_{2}$ = 7500, $\beta_{3}$ = 2.7e-6, $\alpha_{3} = $ 3e-3 , $\lambda = $5e-4, $steps_{3}$ = 23500, $\delta$ = 0.5.

\noindent\textbf{Model Architecture:}
The regression task's architecture has nine fully connected layers with 300 neurons per layer, which is the same as used in \cite{javed2019meta}. ReLu activation function is used for non-linearity.

We use sinusoidal functions as regression tasks. We sample 400 functions for the training and 500 functions for the evaluation. As we can observe from Table~\ref{tab:Table4}, our method consistently outperforms all the baseline methods. 

\begin{table}[t]
    \centering
\scalebox{0.7}{
    \addtolength{\tabcolsep}{0pt}
    \begin{tabular}{cccccc}
        \toprule
         \textbf{Tasks}  &\textbf{Pretrained} &\textbf{SRNN} &\textbf{MRCL} & \textbf{OML} &\textbf{Ours}\\
        \midrule
        
        1  & 0.03 & \textbf{0.01 $\pm$ 0.003} & 0.11 $\pm$ 0.10 & 0.08  & 0.08 $\pm$ 0.06\\
        
        2  & 0.42 & 0.32 $\pm$ 0.27 &0.23 $\pm$ 0.16 & 0.23  &\textbf{0.12 $\pm$ 0.13} \\
        
        3  & 0.73 & 0.41 $\pm$ 0.35 &0.37 $\pm$ 0.18 & 0.26 & \textbf{0.09 $\pm$ 0.05} \\
        
        4  & 0.87 & 0.50 $\pm$ 0.34  &0.28 $\pm$ 0.13 & 0.22  & \textbf{0.16 $\pm$ 0.13}\\
       
        5  & 1.03 & 0.64 $\pm$ 0.33  &0.28 $\pm$ 0.12 & 0.26  & \textbf{0.16 $\pm$ 0.10 }\\
        
        6  & 1.04 & 0.64 $\pm$ 0.40 & 0.34 $\pm$ 0.17 & 0.28  & \textbf{0.17 $\pm$ 0.08} \\
        
        7  & 1.40 & 0.87 $\pm$ 0.57 &0.33 $\pm$ 0.19 & 0.29  & \textbf{0.24 $\pm$ 0.13}\\
        
        8  & 1.22 & 0.76 $\pm$ 0.36 &0.30 $\pm$ 0.16 & 0.27   & \textbf{0.19 $\pm$ 0.10 } \\
        
        9  & 1.33 & 0.80 $\pm$ 0.37 &0.33 $\pm$ 0.18 & 0.32  & \textbf{0.22 $\pm$ 0.10}\\
       
        10  & 1.45 & 0.90 $\pm$ 0.45 &0.40 $\pm$ 0.23 & 0.35  & \textbf{0.22 $\pm$ 0.09 }\\
        \bottomrule
        \end{tabular}}\\
           \caption{MSE loss on 10 regression tasks learned in an incremental fashion averaged over 50 runs. The tasks column refers to loss calculated after having seen the specified number of tasks.} 
    \label{tab:Table4}
  \end{table}

\subsection{Generative Incremental Learning Experiments on MNIST Dataset} 

\noindent\textbf{Implementation Details:}
We use MNIST \cite{lecun-mnisthandwrittendigit-2010} dataset for this experiment. We follow the same algorithm for the generative setting also. We use a conditional GAN for our experiment. The classification loss is replaced by the discriminative and generative loss. Classes 5, 6, 7 are used for testing, and the rest of the classes are used as the training set. For the learning trajectory during training, $\tau_{tr}$ constitutes of 10 samples of a class randomly sampled from the train set, and $\tau_{val}$ consists of 15 samples where 10 samples are randomly sampled from the train set, and the remaining five samples belong to the class used in $\tau_{tr}$. The evaluation protocol for the generative setting is as follows: We learn from classes 5, 6, and 7 sequentially. While learning from a particular class, we do batch updates with batch size 64 for 25 epochs before learning the next class. We have demonstrated the model's ability to prevent catastrophic forgetting by generating samples for class 5 after learning classes 5, 6, 7 in a continual fashion.

\noindent \textbf{Hyperparameter Settings:}
We train our mode using hyperparameters: $\beta_{1}$ = 1e-5, $\alpha_{1} = $ 1e-5, $steps_{1}$ = 300000, $\beta_{3}$ = 1e-5, $\alpha_{3} = $ 1e-5 , $\lambda = $1e-4, $steps_{3}$ = 3500000, $\delta$ = 1.0.

\noindent\textbf{Model Architecture:}
We use conditional-gan with Wasserstein loss~\cite{arjovsky2017wasserstein} for generative setup. The Generator and Discriminator architecture are given as follows.

\textbf{Generator}: Generator is a five-layered fully connected neural network of size (110 $\times$ 128, 128 $\times$ 256, 256 $\times$ 512, 512 $\times$ 1024, 1024 $\times$ 1024). For the non-linearity, Leaky ReLU is used with 0.2 as the negative slope value.

\textbf{Discriminator}: Discriminator is a four-layered fully connected neural network of size (1034 $\times$ 512, 512 $\times$ 512, 512 $\times$ 512, 512 $\times$ 1). For the non-linearity, Leaky ReLU is used with 0.2 as the negative slope value.

\begin{figure}[h]
    \centering
     \includegraphics[scale=0.5]{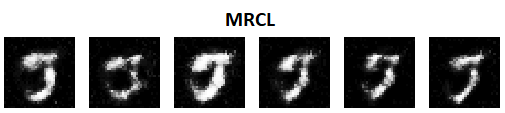}\\
    \includegraphics[scale=0.5]{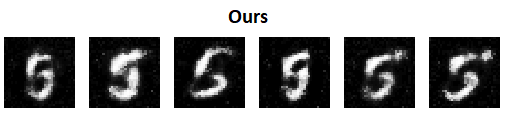}
   \captionof{figure}{Comparison of performance our model against MRCL on generative continual learning tasks. Classes 5 , 6, 7 were learnt sequentially. As we can observe the 5's generated by MRCL have been deformed to 7's as they had been learnt recently, but our model is still able to retain the shape of the 5's which was learnt at the earliest stages.}
\label{fig:gan}
 \end{figure}

We use three classes (say 5, 6, 7) as the test set and the remaining classes are used as the training set. For evaluation, we learn from 1600 examples each of the classes in an online fashion. As we can observe from Figure~\ref{fig:gan}, the quality of images generated by KCCIOL is much better than MRCL. The images generated by MRCL are highly distorted, but images generated by our model are slightly distorted, as it is able to retain the knowledge about previous classes. Therefore the KCCIOL does not outperform the discriminative model only but the generative model also.

\begin{figure}[h]
    \centering
    \includegraphics[scale=0.33]{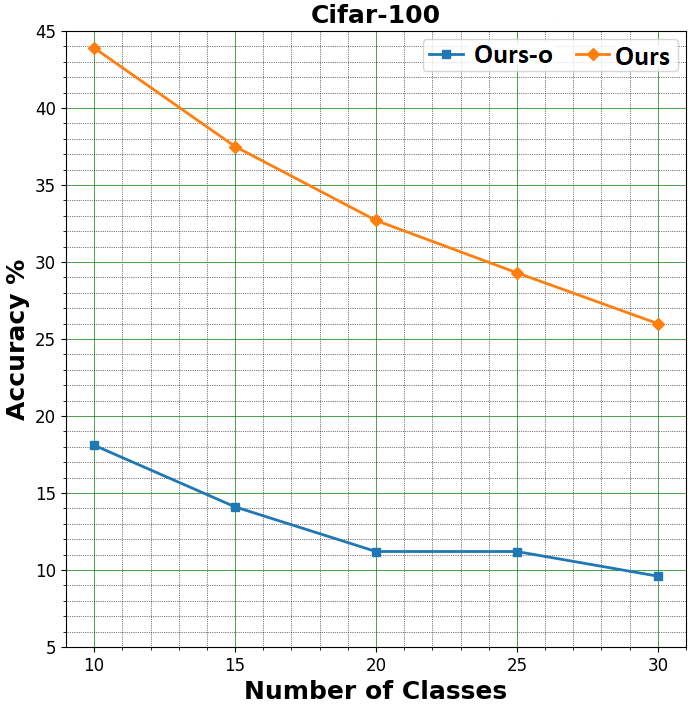}
    \includegraphics[scale=0.33]{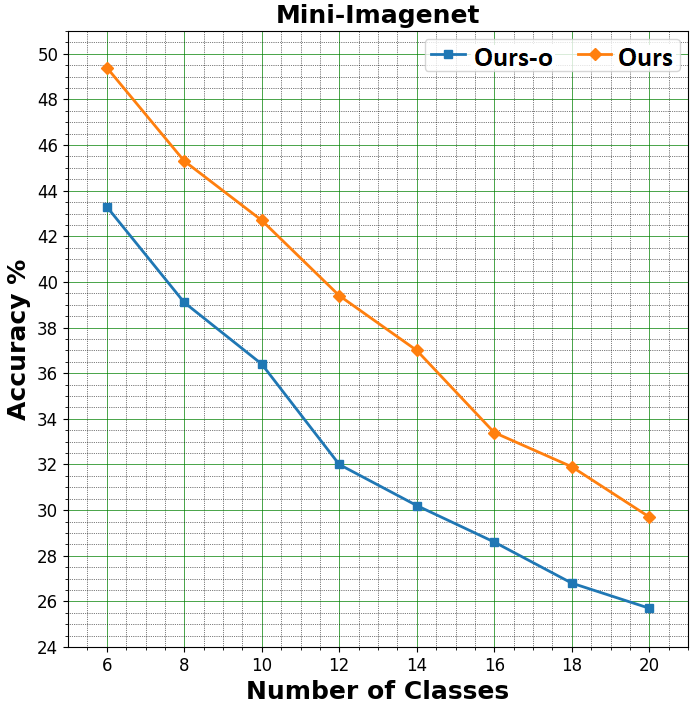}
                   \captionof{figure}{Comparison of performance of KCCIOL and KCCIOL-o (without knowledge transfer) on continual learning tasks. The x axis indicates the number of classes  learnt continually averaged over 50 test tasks on a) CIFAR-100 dataset, b) Mini-Imagenet dataset.}
\label{fig:withoutknowledge}
\end{figure}                   

\begin{figure}[h]
  \centering
  \subfigure [Weight distribution] {\includegraphics[scale=.32]{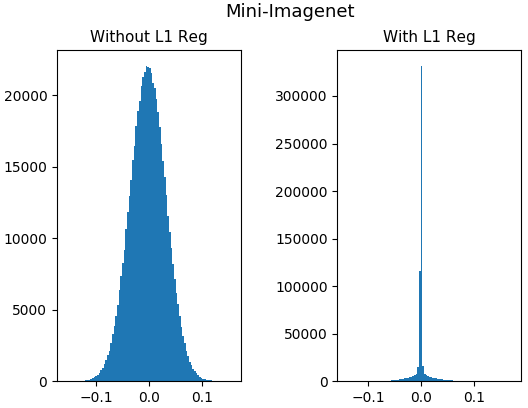}}
  \subfigure [Acc. vs. Masking level]
  {\includegraphics[height=3.5cm,width=3.8cm]{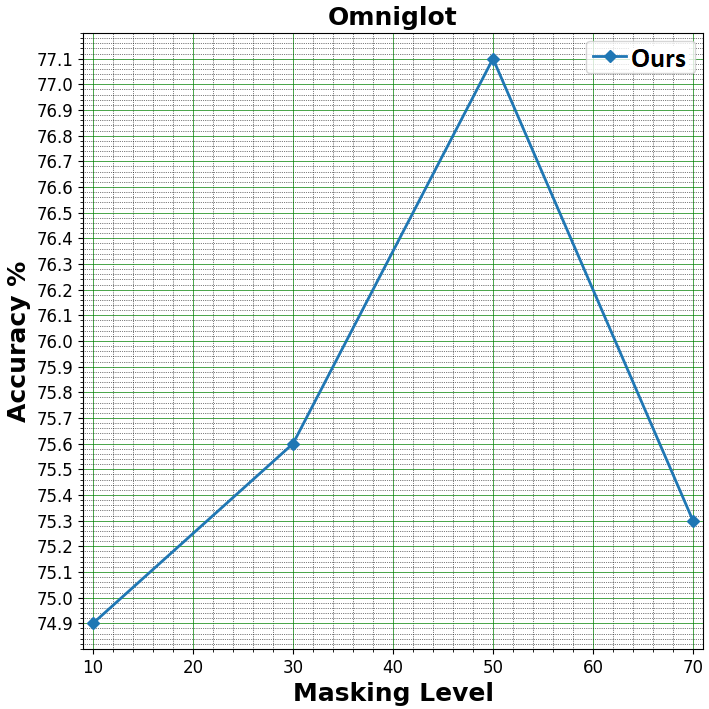}}
  \captionof{figure}{(a) Histogram of 5th convolution layer with and without $l_1$ fine-tuning on Mini-Imagenet dataset. We can observe from the flat tail of the distribution that the knowledge is being transferred to a subset of parameters. (b) Comparison of performance of KCCIOL as a function of masking level for the omniglot dataset. }
  \label{fig:knowledgetransfer_masking}
\end{figure}

\begin{figure}[h]
    \centering
     \includegraphics[scale=0.26]{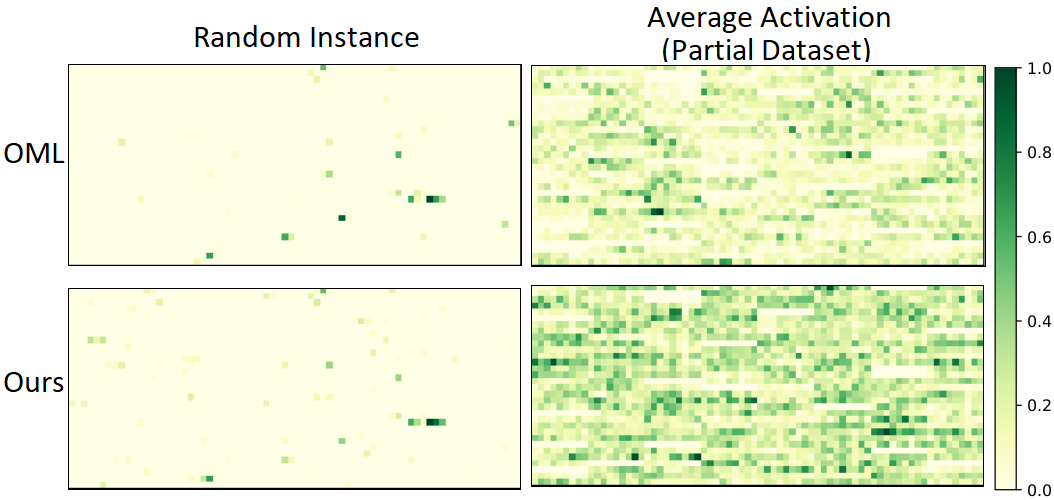}
     \vspace{-0.5em}
     \includegraphics[height=2.2cm,width=3.6cm]{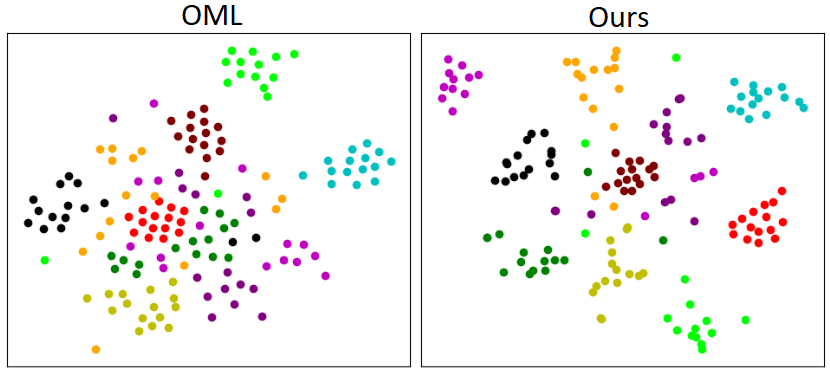}
\caption{We reshape the 2304 length representation vectors into 32x72, normalize them to have a maximum value of one, and visualize them. The figure in column-1 shows activation for a random instance. The figure in column-2 shows average activation on 10\% of the meta test set of the omniglot dataset. Columns-3,4 show the TSNE plots of representation learned by OML and KCCIOL on 10 randomly selected classes from the meta testing set of the omniglot.}
\label{fig:heatmap}
\end{figure}

\subsection{Ablation Study and Analysis}
\textbf{Parameter Importance:} 
We plot the accuracy as a function of the number of classes learned continually on the CIFAR-100 and Mini-Imagenet dataset as shown in Figure~\ref{fig:withoutknowledge} for KCCIOL and KCCIOL-o (without knowledge transfer). In the case of KCCIOL, only the important parameters are constrained. Whereas in KCCIOL-o, all the parameters are constrained. To apply a constraint loss on the basis of importance, we first perform a knowledge transfer step to transfer all the knowledge to the important parameters. We can infer from the flat tail in Figure~\ref{fig:knowledgetransfer_masking} (a) that the knowledge is indeed transferred to a subset of the model parameters. We can observe from Figure~\ref{fig:withoutknowledge} that constraining only important parameters (KCCIOL) instead of constraining all the parameters (KCCIOL-o) boosts the performance significantly on continual learning tasks. Therefore, weight constraint on important parameters substantially improves performance over a different number of classes.

\noindent\textbf{Masking level:}
Masking level is defined as the fraction of model parameters constrained during training. We plot the accuracy as a function of the masking level on the Omniglot dataset. We can observe from Figure~\ref{fig:knowledgetransfer_masking} (b) that we get the best performance at 50\% masking level, i.e., we apply weight constraint over 50 percent of the model parameters (important parameters) based on the absolute value of weights.  If we choose the masking level below 50\% then we may lose accuracy because we are not applying weight constraint over all important parameters. If we choose a masking level above 50\% then we may lose accuracy because we have not rejuvenated all the unimportant parameters.

\noindent\textbf{Analysis of Representation Learned by KCCIOL:}
Figure \ref{fig:heatmap} depicts the activation maps of the feature representation generated by the models. As we can observe for a random image instance of meta-test of the omniglot dataset, the activation map of both OML and KCCIOL are sparse. The strength of average activation is much better in KCCIOL as compared to OML.  This indicates that the model produced by OML is not being used to its full capacity. In KCCIOL, we enhance the model's performance as shown in  Figure \ref{fig:heatmap}  where, on average, a significant proportion of activations is strong for KCCIOL as compared to OML. Columns-3, 4 in Figure \ref{fig:heatmap} show that the representation learned by our method (KCCIOL) is better than OML.

\end{document}